# A Harmonic Mean Linear Discriminant Analysis for Robust Image Classification

Shuai Zheng, Feiping Nie, Chris Ding, Heng Huang
*Department of Computer Science and Engineering,*
*University of Texas at Arlington, TX, USA*
zhengs123@gmail.com, feipingnie@gmail.com, chqding@uta.edu, heng@uta.edu

*Abstract*—Linear Discriminant Analysis (LDA) is a widely-used supervised dimensionality reduction method in computer vision and pattern recognition. In null space based LDA (NLDA), a well-known LDA extension, between-class distance is maximized in the null space of the within-class scatter matrix. However, there are some limitations in NLDA. Firstly, for many data sets, null space of within-class scatter matrix does not exist, thus NLDA is not applicable to those datasets. Secondly, NLDA uses *arithmetic mean* of between-class distances and gives equal consideration to all between-class distances, which makes larger between-class distances can dominate the result and thus limits the performance of NLDA. In this paper, we propose a harmonic mean based Linear Discriminant Analysis, Multi-Class Discriminant Analysis (MCDA), for image classification, which minimizes the reciprocal of weighted *harmonic mean* of pairwise between-class distance. More importantly, MCDA gives higher priority to maximize small between-class distances. MCDA can be extended to multi-label dimension reduction. Results on 7 single-label data sets and 4 multi-label data sets show that MCDA has consistently better performance than 10 other single-label approaches and 4 other multi-label approaches in terms of classification accuracy, macro and micro average F1 score.

*Keywords*-Dimensionality Reduction; Linear Discriminant Analysis; Image Classification

## I. INTRODUCTION

Researchers and engineers nowadays have larger and larger data with very high dimension to be processed everyday [1]. For example, in image classification, a small image of size $100 \times 100$ will have $10,000$ dimension. In biology science, high-dimensional gene expression data is used to predict tumor [2]. However, there is always an underlying low-dimensional structure which can capture the underlying main attributes of high-dimensional data. What's more, high-dimensional data costs a lot of computing resources. Dimensionality reduction algorithms have been proposed to extract important features from high-dimensional data.

Dimensionality reduction is important in many applications of statistics, pattern recognition and machine learning. Many methods have been proposed for dimensionality reduction, such as principal component analysis (PCA) [3] and linear discriminant analysis (LDA) [4]. LDA is a popular supervised dimensionality reduction. To be specific, let $X \in \Re^{p \times n}$ be the data matrix, and $X = (\mathbf{x}_1, \cdots, \mathbf{x}_n)$, where $p$ is data dimension, $n$ is number of data points. Let $G \in \Re^{p \times k}$ be the transformation matrix to a $k$-dimensional subspace. The between-class scatter matrix $S_b$, within-class scatter matrix $S_w$ and total scatter matrix $S_t$ is defined as:

$$S_b = \sum_{k=1}^{K} n_k (\mathbf{m}_k - \mathbf{m})(\mathbf{m}_k - \mathbf{m})^T, \quad (1)$$

$$S_w = \sum_{k=1}^{K} \sum_{\mathbf{x}_i \in k} (\mathbf{x}_i - \mathbf{m}_k)(\mathbf{x}_i - \mathbf{m}_k)^T, \quad (2)$$

$$S_t = S_b + S_w, \quad (3)$$

where $K$ is total class number, $n_k$ is number of points in class $k$, $\mathbf{m}_k = \sum_{\mathbf{x}_i \in k} \mathbf{x}_i / n_k$ is the mean of class $k$, $\mathbf{m} = \sum_{i=1}^{n} \mathbf{x}_i / n$ is the mean of entire data set. $S_b$, $S_w$ and $S_t$ are semi-positive definite matrices. Classical LDA finds a transformation matrix $G$ by solving the problem:

$$\max_G \mathrm{Tr} \frac{G^T S_b G}{G^T S_w G}. \quad (4)$$

When $G^T S_w G = 0$, i.e., there exist null space of $S_w$, the above formulation has some difficulty. In null space based LDA (NLDA) [5], the between-class distance is maximized in the null space of within-class scatter matrix $S_w$,

$$\max_G \mathrm{Tr}(G^T S_b G), \text{ s.t. } G^T S_w G = \mathbf{0}, G^T G = I \quad (5)$$

which is based on the idea that the null space of $S_w$ contains sufficient discriminant information. The dimension of the null space of $S_w$ is at least $p - (n - K)$. When $p \leq (n - K)$, the null space of $S_w$ may not exist. In the era of big data, we usually have sufficient amount of training data, so $p$ is usually less than $(n - K)$. Thus, NLDA is not applicable to many problems. What's more, NLDA gives equal consideration to all between-class distances, which makes larger between-class distances could dominate the objective function and thus limits the performance of NLDA.

In this paper, we propose a harmonic mean based Linear Discriminant Analysis, Multi-Class Discriminant Analysis (MCDA), to overcome the limitations of NLDA. MCDA minimizes within-class distance and maximizes weighted pairwise between-class distance. More importantly, MCDA

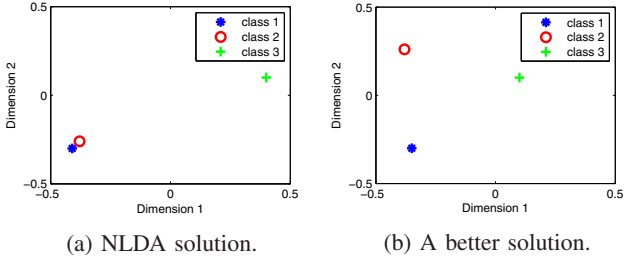

(a) NLDA solution.   (b) A better solution.

Figure 1: Limitation of NLDA: (a) NLDA solution: $\text{Tr}(G^T S_b G) = 0.6401$, $G^T S_w G = \mathbf{0}$; (b) A better solution: $\text{Tr}(G^T S_b G) = 0.4445$, $G^T S_w G = \mathbf{0}$.

gives higher priority to maximize small pairwise between-class distance.

## II. LIMITATIONS OF NLDA

NLDA has some limitations. Firstly, null space of within-class scatter matrix $S_w \in \Re^{p \times p}$ may not exist for many data sets, where $p$ is data dimension. Rank of $S_w$ is at most $n-K$, where $n$ is sample number, $K$ is class number. Thus the null space dimension of $S_w$ is at least $p-(n-K)$. When $p \leq (n-K)$, the null space of $S_w$ may not exist.

Secondly, NLDA gives equal consideration to all between-class distances, which makes larger between-class distances could dominate the objective function and thus limits the performance of NLDA. NLDA solves the problem of Eq.(5), which maximizes the distance between $\mathbf{m}_k$ (the center of class $k$) and $\mathbf{m}$ (the center of all data). However, it gives equal consideration to all between-class distances. Figure 1 shows two solutions on a toy data. This toy data has 3 classes and each class contains 10 points. Figure 1a shows the NLDA solution, where $G^T S_w G = \mathbf{0}$, ensures the solution is in the null space of within-class scatter matrix $S_w$. Data points of the same class overlap with each other. The maximized sum of squared between-class distance $\text{Tr}(G^T S_b G) = 0.6401$. However, as we can see from the figure, even though the sum of squared between-class distance is maximized, class 1 and class 3 are not discriminated. Figure 1b gives a better solution, where all the 3 classes are separated evenly. Solution of Figure 1b is also in the null space of within-class scatter matrix $S_w$, but the sum of squared between-class distance $\text{Tr}(G^T S_b G) = 0.4445$, which is not maximized from the view of NLDA.

## III. A HARMONIC MEAN BETWEEN-CLASS DISTANCE OBJECTIVE

As we can see from the demonstration in Figure 1, pairwise between-class distance plays an important role in the result. Figure 1b is a better solution than Figure 1a, because all 3 classes in the solution are clearly separated and no two classes are too close to each other. In order to achieve this goal, we introduce the use of pair-wise between-class distance. To incorporate pairwise between-class distance into our objective, we define pairwise between-class scatter matrix $B_{k_1 k_2}$ for class $k_1$ and $k_2$ as:

$$B_{k_1 k_2} = (\mathbf{m}_{k_1} - \mathbf{m}_{k_2})(\mathbf{m}_{k_1} - \mathbf{m}_{k_2})^T. \quad (6)$$

We wish to maximize the between class distances. Instead of the traditional approach of $\max \text{Tr}(G^T S_b G)$ in Eq.(5), we maximize all pairs of the pairwise between-class distance:

$$\max_G \sum_{k_1=1}^{K-1} \sum_{k_2=k_1+1}^{K} n_{k_1} n_{k_2} \text{Tr}(G^T B_{k_1 k_2} G), \text{ s.t. } G^T G = I \quad (7)$$

where $n_{k_1}$ is the number of points in class $k_1$, $n_{k_2}$ is the number of points in class $k_2$. The key point of this new approach is that $S_b$ is a global summation, insensitive to individual class variations, but $B_{k_1 k_2}$ is more sensitive to individual class variations.

The weight $n_{k_1} n_{k_2}$ controls the relative importance of between class distance between class $k_1$ and $k_2$. The reason why we use $n_{k_1} n_{k_2}$ is because the importance of separating two classes is based on the number of points in these two classes respectively.

In above approach, although $B_{k_1 k_2}$ is more sensitive than $S_b$, the addition of all pairs, or the *arithmetic mean* of pairwise between-class distance, is not robust: one large between-class distances could dominate the objective function.

It is well-known that the *harmonic mean* is more robust than *arithmetic mean*. Thus, we propose to minimize the inverse of *harmonic mean* of pairwise between-class distance:

$$\min_G \sum_{k_1=1}^{K-1} \sum_{k_2=k_1+1}^{K} \frac{n_{k_1} n_{k_2}}{\text{Tr}(G^T B_{k_1 k_2} G)} \text{ s.t. } G^T G = I. \quad (8)$$

Clearly, the difficult case is when two classes are close in feature space; In this case, their between-class distances is small. The objective of Eq.(8) gives higher weight to this pair of classes, therefore correctly emphasize is critical part of the discrimination task. In contrast, the objective of Eq.(7) gives small weight to this difficult pair, thus incorrectly de-emphasize the critical part of the discrimination task. This is the key reason we propose the *harmonic mean* objective.

## IV. MULTI-CLASS DISCRIMINANT ANALYSIS (MCDA)

In this section, we propose a generalized, efficient and stable LDA formulation, Multi-Class Discriminant Analysis (MCDA).

Since in NLDA, for many applications, the null space of within-class scatter matrix does not exist, which means there is no such transformation matrix $G$, such that $G^T S_w G = \mathbf{0}$, or $\text{Tr}(G^T S_w G) = 0$. A revised version of the objective is to minimize $\text{Tr}(G^T S_w G)$, so we have:

$$\min_G \text{Tr}(G^T S_w G) \text{ s.t. } G^T G = I. \quad (9)$$

**Algorithm 1** Gradient descent algorithm for MCDA.

**Input:** Data matrix $X \in \Re^{p \times n}$ with $n$ data points and $p$ dimension; class indicator matrix $Y \in \Re^{n \times K}$, $K$ is number of classes; subspace dimension $k$
**Output:** Projection matrix $G \in \Re^{p \times k}$
1: Initialize $G$
2: Compute $S_w$ and $B_{k_1 k_2}$ using Eq.(2) and Eq.(6)
3: **while** Objective value Eq.(10) dose not converge **do**
4:     Compute gradient using Eq.(11)
5:     Update $G$ using gradient
6:     Use SVD to enforce constraint $G^T G = I$ (every few iterations)
7: **end while**

In MCDA, we use Eq.(8) to maximize between-class distance, because Eq.(8) is a robust between-class distance objective and it gives higher priority to maximize small between-class distances. To summarize, the objective of Multi-Class Discriminant Analysis (MCDA) is proposed as

$$\min_G J_{MCDA} = \gamma \text{Tr}(G^T S_w G) + \sum_{k_1=1}^{K-1} \sum_{k_2=k_1+1}^{K} \frac{n_{k_1} n_{k_2}}{\text{Tr}(G^T B_{k_1 k_2} G)},$$
$$\text{s.t.} \quad G^T G = I, \quad (10)$$

where $\gamma$ controls the weight between minimizing within-class distance and maximizing between-class distance, constraint $G^T G = I$ ensures the columns of solution $G$ are linearly independent. When $\gamma \to \infty$, Eq.(10) focuses on minimizing within-class distance only, which is equal to finding the null space of within-class scatter matrix $S_w$. When $\gamma \to 0$, Eq.(10) focuses on maximizing pairwise between-class distance.

## V. ALGORITHM

We use gradient descent to solve Eq.(10). The gradient is:

$$\frac{\partial J_{MCDA}}{\partial G} = 2\gamma S_w G - \sum_{k_1=1}^{K-1} \sum_{k_2=k_1+1}^{K} \frac{2 n_{k_1} n_{k_2} B_{k_1 k_2} G}{(\text{Tr} G^T B_{k_1 k_2} G)^2}. \quad (11)$$

In order to enforce the constraint requirement $G^T G = I$, we use SVD decomposition to make $G$ orthonormal every few iterations. If we apply SVD in every iteration, it will be computational expensive. Thus, we can apply SVD every few iterations, which is on projection matrix G (subspace dimension is very small), and is very fast. Algorithm 1 summarizes the steps to solve Eq.(10). The objective is optimized in an iterative fashion. There is no need to do Eigen decomposition or matrix inverse for scatter matrices.

When initializing matrix $G$, if subspace dimension $k <= K-1$, we can use classical LDA Eq.(4) solution to initialize $G$; when $k > K-1$, we can use trace ratio LDA Eq.(12) solution to initialize $G$. This ensures that our approach can find a better solution than other LDA formulations (see experiments part for comparison).

Table I: Single-label dataset overview.

| Dataset | sample # $n$ | dimension $p$ | class # $K$ |
|---|---|---|---|
| Caltech07-HOG | 210 | 432 | 7 |
| Caltech20-HOG | 1230 | 432 | 20 |
| MSRC-HOG | 210 | 432 | 7 |
| MSRC-GIST | 210 | 512 | 7 |
| ATT | 400 | 644 | 40 |
| BinAlpha | 1014 | 320 | 26 |
| MNIST | 150 | 784 | 10 |

## VI. ILLUSTRATION

To show the effectiveness of proposed MCDA, Figure 2 visualizes two real data sets, Caltech and MSRC, in 2-D subspace using PCA, LDA, null space LDA (NLDA) and MCDA. In this example, we take 4 classes from Caltech101 and MSRC, 30 data points in each class. We extract 432-dimensional HOG feature. Figure 2a and 2e show the Caltech and MSRC data projected in 2-D PCA subspace. As we can see, data points from 4 classes are mixed together. Figure 2b and 2f show the data projected in 2-D LDA subspace. Figure 2c and 2g shows the Caltech and MSRC data projected in 2-D NLDA subspace. Data points have been projected on 4 different data points, each of which includes 30 overlapped points, which is because the within-class distance in this subspace is 0 now. However, class 1 and class 4 in both Figure 2c and 2g are still very close. This is due to the limitations of NLDA we discussed in Figure 1. By using MCDA proposed in Eq.(10), there are no two classes becoming too close in Figure 2d and 2h. MCDA takes weighted harmonic mean of pairwise between-class distance into account and gives higher priority to small between-class distances.

## VII. CONNECTION WITH TRACE RATIO

Trace ratio was proposed in [6], [7], [8] to solve the following problem:

$$\max_G \frac{\text{Tr}(G^T S_b G)}{\text{Tr}(G^T S_t G)} \quad \text{s.t.} \quad G^T G = I. \quad (12)$$

Since Eq.(12) maximize $\text{Tr}(G^T S_b G)$, it maximizes the arithmetic mean of between-class distance as well and also suffers from the robustness problem as discussed in Eq.(7).

Trace ratio problem can be reduced to NLDA when the reduced subspace dimension $k$ is not larger than the dimension of null space of $S_w$. Using $S_t = S_b + S_w$, Eq.(12) is equivalent to $\max_G \frac{\text{Tr}(G^T S_b G)}{\text{Tr}(G^T S_b G) + \text{Tr}(G^T S_w G)} = \max_G \frac{1}{1 + \text{Tr}(G^T S_w G)/\text{Tr}(G^T S_b G)}$. Since $S_b$, $S_w$ and $S_t$ are all semi-positive definite, the maximum objective value can be achieved when $\text{Tr}(G^T S_w G) = 0$. This means that the optimal solution $G$ is in the null space of within-class scatter matrix $S_w$.

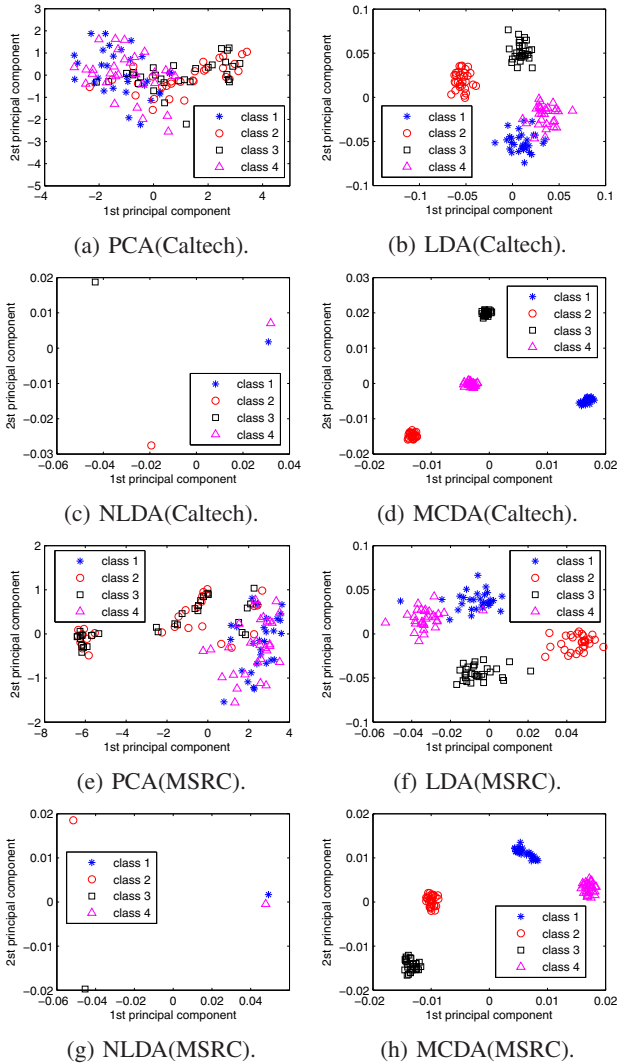

(a) PCA(Caltech). (b) LDA(Caltech).
(c) NLDA(Caltech). (d) MCDA(Caltech).
(e) PCA(MSRC). (f) LDA(MSRC).
(g) NLDA(MSRC). (h) MCDA(MSRC).

Figure 2: Visualization of Caltech and MSRC data in 2-D subspace.

Table II: Multi-label dataset overview.

| Data | sample # $n$ | dimension $p$ | class # $K$ |
|---|---|---|---|
| Barcelona | 139 | 48 | 4 |
| Yeast | 2,417 | 103 | 14 |
| MSRC-SIFT | 591 | 240 | 23 |
| Scene | 2,407 | 294 | 6 |

## VIII. MULTI-LABEL MCDA

In image and video annotation, each image is usually associated with several different conceptual classes. Let's take two sample images from MSRC data in Figure 4 as an example. Figure 4a is annotated using 3 words: sky, plane and grass; Figure 4b is annotated using 3 words: car, building, road. In machine learning, such problem that requires each data point to be assigned to multiple different classes

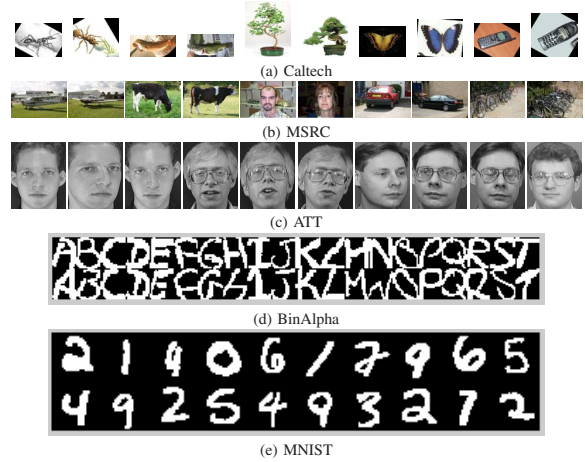

(a) Caltech (b) MSRC (c) ATT (d) BinAlpha (e) MNIST

Figure 3: Example images.

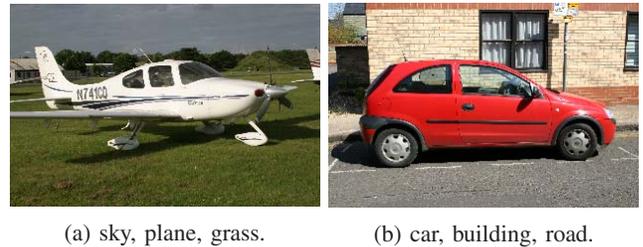

(a) sky, plane, grass. (b) car, building, road.

Figure 4: Sample images from MSRC data set. Each image is annotated with several different words. In a multi-label multi-class classification problem, each image is classified into more than 1 class.

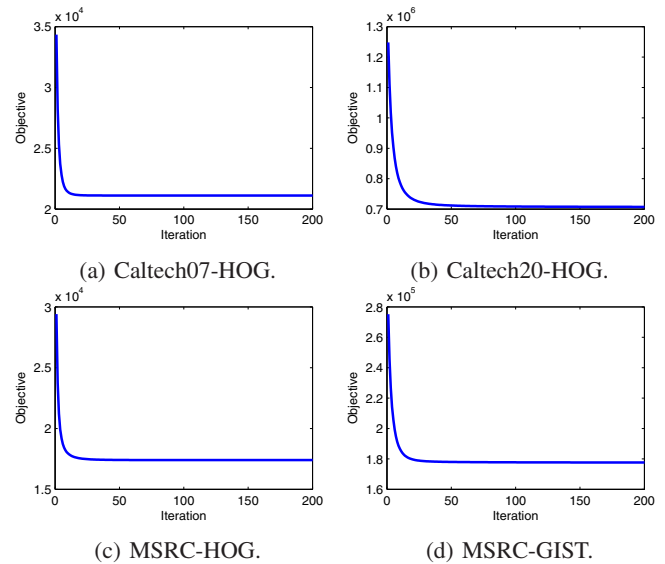

(a) Caltech07-HOG. (b) Caltech20-HOG.
(c) MSRC-HOG. (d) MSRC-GIST.

Figure 5: Objective function converges in about 50 iterations using Algorithm 1.

is called multi-label classification problem. In contrast, in

Table III: Classification Accuracy on 7 data sets(subspace dimension is $K-1$, except NLDA, best results are in bold).

| Dataset | MCDA | LDA | NLDA | TraceRatio | sdpLDA | MMC | RLDA | ULDA | OLDA | OCM | OLSLDA |
|---|---|---|---|---|---|---|---|---|---|---|---|
| Caltech07-HOG | **0.8048** | 0.7857 | 0.6762 | 0.6762 | 0.7619 | 0.7905 | 0.7619 | 0.7619 | 0.8038 | 0.7619 | 0.7381 |
| Caltech20-HOG | **0.7009** | 0.6872 | — | 0.4553 | 0.6815 | 0.5941 | 0.6815 | 0.6693 | 0.6841 | 0.6815 | 0.5589 |
| MSRC-HOG | **0.7857** | 0.7810 | 0.5714 | 0.5714 | 0.7286 | 0.7714 | 0.7286 | 0.7381 | 0.7571 | 0.7286 | 0.5238 |
| MSRC-GIST | **0.8524** | 0.8143 | 0.6286 | 0.6286 | 0.8190 | 0.8333 | 0.8190 | 0.8190 | 0.8190 | 0.8190 | 0.6095 |
| ATT | **0.9750** | 0.9675 | 0.9675 | 0.8675 | 0.9625 | 0.9675 | 0.9625 | 0.9625 | 0.9500 | 0.9625 | 0.9650 |
| BinAlpha | **0.8330** | 0.8228 | — | 0.4638 | 0.8176 | 0.7720 | 0.8150 | 0.8150 | 0.7808 | 0.8176 | 0.5371 |
| MNIST | **0.8867** | **0.8867** | 0.8733 | 0.7400 | 0.8467 | 0.8667 | 0.8467 | 0.8533 | 0.8600 | 0.8467 | 0.8667 |

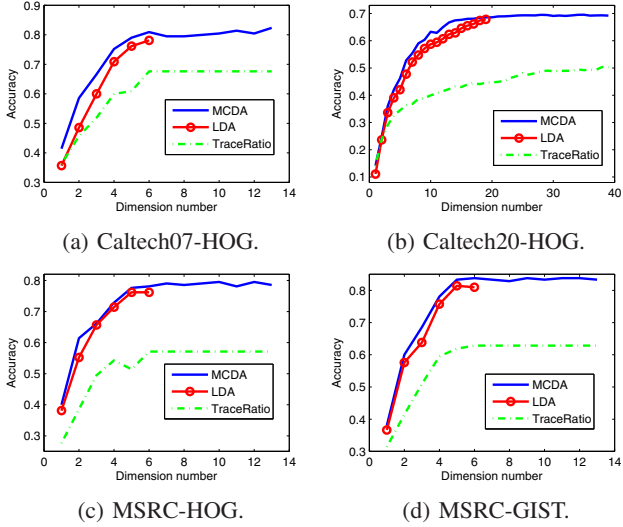

Figure 6: Effect of reduced dimension number.

(a) Caltech07-HOG. (b) Caltech20-HOG. (c) MSRC-HOG. (d) MSRC-GIST.

traditional single-label classification, which is also called single-label multi-class classification, each data point is only classified into one category. Multi-label multi-class problem is more generalized than single-label multi-class problem.

An important difference between single-label classification and multi-label classification is that class memberships in single-label classification are mutually exclusive, while class memberships in multi-label classification are overlapped with 2 or more classes. Class memberships can be inferred from label correlations, which can be used to improve classification. It has stimulated many multi-label learning algorithms [9] [10] [11] [12].

However, Linear Discriminant Analysis (LDA) by nature is derived for single-label classification. Wang proposed a multi-label formulation of scatter matrices for multi-label data in [12]. Multi-label class indicator matrix $Y \in \Re^{n \times K}$ is defined as

$$Y_{ik} = \begin{cases} 1, & \text{if point } i \text{ is in class } k. \\ 0, & \text{otherwise.} \end{cases} \quad (13)$$

For data point $i$, $\sum_k Y_{ik} > 1$, which means that data $i$ belongs to more than 1 class. Multi-label between-class scatter matrix $\widetilde{S_b}$ and within-class scatter matrix $\widetilde{S_w}$ are defined as follows [12]:

$$\widetilde{S_b} = \sum_{k=1}^{K}(\sum_{i=1}^{n} Y_{ik})(\mathbf{m}_k - \mathbf{m})(\mathbf{m}_k - \mathbf{m})^T, \quad (14)$$

$$\widetilde{S_w} = \sum_{k=1}^{K}\sum_{i=1}^{n} Y_{ik}(\mathbf{x}_i - \mathbf{m}_k)(\mathbf{x}_i - \mathbf{m}_k)^T, \quad (15)$$

where $\mathbf{m}_k$ is the mean of class $k$ and $\mathbf{m}$ is global mean, defined as follows:

$$\mathbf{m}_k = \frac{\sum_{i=1}^{n} Y_{ik}\mathbf{x}_i}{\sum_{i=1}^{n} Y_{ik}}, \quad \mathbf{m} = \frac{\sum_{k=1}^{K}\sum_{i=1}^{n} Y_{ik}\mathbf{x}_i}{\sum_{k=1}^{K}\sum_{i=1}^{n} Y_{ik}}. \quad (16)$$

Eq.(14,15) is also equivalent to Eq.(28, 29, 30) in [13]. Using Eq.(15) and Eq.(16), the objective of Multi-label MCDA is proposed as

$$\min_{G} \gamma \text{Tr}(G^T \widetilde{S_w} G) + \sum_{k_1=1}^{K-1}\sum_{k_2=k_1+1}^{K} \frac{n_{k_1} n_{k_2}}{\text{Tr}(G^T \widetilde{B}_{k_1 k_2} G)}, \quad (17)$$
$$\text{s.t.} \quad G^T G = I,$$

where $\widetilde{B}_{k_1 k_2}$ is between-class scatter matrix for class $k_1$ and $k_2$:

$$\widetilde{B}_{k_1 k_2} = (\mathbf{m}_{k_1} - \mathbf{m}_{k_2})(\mathbf{m}_{k_1} - \mathbf{m}_{k_2})^T. \quad (18)$$

$$n_{k_1} = \sum_{i=1}^{n} Y_{ik_1}, \quad n_{k_2} = \sum_{i=1}^{n} Y_{ik_2}. \quad (19)$$

## IX. EXPERIMENTS

In this section, we first study the convergence of Algorithm 1, and the effect of reduced dimension number of MCDA. Then we compare classification accuracy of MCDA with 10 other subspace learning algorithms using reduced dimension number $K-1$ ($K$ is class number). Finally, we experiment the classification accuracy and macro and micro average F score for multi-label data sets.

**Single-label Dataset** 7 single-label datasets are used in this experiment. Caltech101[14] contains 101 object categories. We then use VLFeat [15] to extract HOG feature. Caltech07-HOG contains 7 categories randomly selected from Caltech101 and each category has 30 images. Caltech20-HOG contains 20 categories randomly selected from Caltech101 and each category has 30 images. MSRC [16] is from MSRC data base v1 and contains 7 classes with 30 images in each class. We use the HOG and GIST feature of MSRC. Other datasets include face datasets ATT

[17], digit datasets MNIST [18] and handwritten alphabets Binalpha. Table I summarizes the attributes of single-label datasets. Figure 3 shows sample images from the data.

**Multi-label Dataset** 4 multi-label datasets are used in this experiment. Barcelona data set contains 139 images with 4 categories, i.e., building, flora, people and sky. Each image has at least two labels. Yeast is a multi-label data from [19]. MSRC [16] is MSRC multi-label data base v2 provided by Microsoft Research Cambridge, which has 591 images annotated by 23 classes. Scene is a multi-label image data from [20]. Table II summarizes the attributes of those datasets.

### A. Convergence of Algorithm 1

We take the first 4 single-label datasets, Caltech07-HOG, Caltech20-HOG, MSRC-HOG, MSRC-GIST, as examples to check the convergence of the Algorithm 1. In order to find a reasonable guess for $\gamma$, the first part $\gamma \text{Tr}(G^T S_w G)$ of Eq.(10) and the second part $\sum_{k_1=1}^{K-1} \sum_{k_2=k_1+1}^{K} \frac{n_{k_1} n_{k_2}}{\text{Tr}(G^T B_{k_1 k_2} G)}$ should be in similar scale. To get an approximate value, we set $G = I$ in Eq.(10), if $\gamma \text{Tr}(G^T S_w G) = \sum_{k_1=1}^{K-1} \sum_{k_2=k_1+1}^{K} \frac{n_{k_1} n_{k_2}}{\text{Tr}(G^T B_{k_1 k_2} G)}$, we have

$$\gamma = \frac{1}{\text{Tr} S_w} \sum_{k_1=1}^{K-1} \sum_{k_2=k_1+1}^{K} \frac{n_{k_1} n_{k_2}}{\text{Tr} B_{k_1 k_2}}.$$

Figure 5 shows the objective value of Eq.(10) while using the above $\gamma$. We can see that all the 4 objective values converge quickly in about 50 iterations.

### B. Effect of subspace dimension

Standard LDA can find subspace dimension from 1 to $K-1$. MCDA does not have dimension limit. So in this part, we study MCDA subspace classification performance with respect to subspace dimension and we compare the performance with standard LDA and trace ratio. For standard LDA, we only compute reduced dimension from 1 to $K-1$. After using dimension reduction, KNN classifier ($knn = 3$) is applied to perform classification. The classification accuracy is the average of 5-fold cross validation results. Figure 6 shows the classification accuracy of MCDA, LDA and Trace Ratio. We can see from the result that MCDA has higher classification accuracy than LDA and Trace Ratio when using the same number of dimension on all 4 data sets.

### C. Single-label classification experiment

On 7 single-label dataset, we compare MCDA with 10 other different methods, including LDA, null space LDA(NLDA) [5], Trace Ratio LDA(TraceRatio) [6], Semi Definite Positive LDA(spdLDA) [21], Maximum Margin Criteria(MMC) [22], regularized LDA(RLDA) [23], Uncorrelated LDA(ULDA) [24], Orthogonal LDA (OLDA) [24],

Table IV: Multi-label classification accuracy (best results are in bold).

| Dataset | MCDA | MLSI | MDDM | MLLS | MLDA |
|---|---|---|---|---|---|
| Barcelona | **0.6745** | 0.6436 | 0.6470 | 0.6524 | 0.6290 |
| Yeast | **0.7386** | 0.7317 | 0.7371 | 0.7364 | 0.7368 |
| MSRC | **0.8860** | 0.8762 | 0.8800 | 0.8807 | 0.8858 |
| Scene | **0.8806** | 0.8534 | 0.8713 | 0.8229 | 0.8771 |

Table V: Multi-label Macro F1 score (best results are in bold).

| Dataset | MCDA | MLSI | MDDM | MLLS | MLDA |
|---|---|---|---|---|---|
| Barcelona | **0.7509** | 0.7286 | 0.7301 | 0.7341 | 0.7169 |
| Yeast | **0.5717** | 0.5568 | 0.5696 | 0.5691 | 0.5693 |
| MSRC | **0.4776** | 0.4334 | 0.4522 | 0.4544 | 0.4773 |
| Scene | **0.6670** | 0.5911 | 0.6411 | 0.5048 | 0.6568 |

Table VI: Multi-label Micro F1 score (best results are in bold).

| Dataset | MCDA | MLSI | MDDM | MLLS | MLDA |
|---|---|---|---|---|---|
| Barcelona | **0.7173** | 0.6891 | 0.6861 | 0.6904 | 0.6772 |
| Yeast | 0.4171 | 0.4026 | 0.4205 | **0.4216** | 0.4213 |
| MSRC | **0.3969** | 0.3510 | 0.3637 | 0.3667 | 0.3959 |
| Scene | **0.6772** | 0.6006 | 0.6493 | 0.5062 | 0.6643 |

Orthogonal Centroid Method(OCM) [25] and Orthogonal Least Squares LDA(OLSLDA)[26].

In experiment, we reduce raw data to $K-1$ dimension. We use 5-fold cross validation to select training and testing data. After selecting training and testing data, we tune parameters based on the selected training data only. We tune weight parameter $\gamma$ in Eq.(10) from $10^{-10}$ to $10^{10}$ and use the best result for the selected training data only. Solution projection matrix $G$ was solved based on only the training set. Then we apply $G$ on both training and testing data and then KNN is applied to do the classification. In our experiment, we set the nearest neighbors number $knn = 3$. The final classification accuracy is the average of 5-fold cross validation results, and is reported in Table III. MCDA outperforms all other 10 algorithms. Note, parameters of other algorithms have also been tuned to the best value, such as the regularization parameter of regularized LDA. For Caltech20-HOG and BinAlpha, null space of $S_w$ does not exist.

### D. Multi-label classification experiment

We compare the performance of Multi-label MCDA with 4 other multi-label algorithms on 4 multi-label datasets in terms of macro accuracy (Table IV), macro-averaged F1-score (Table V) and micro-averaged F1-score (Table VI). F1-score is defined as: $F1 = 2 \times \frac{\text{precision} \times \text{recall}}{\text{precision} + \text{recall}}$. Macro-average is the average based on the overall testing dataset, while micro-average is the average which gives equal weight to each class. Macro-averaged and micro-averaged F1-score are widely used as a metric to evaluate classification performance [27].

The algorithms for multi-label dataset we compared in this section include Multi-label informed Latent Semantic Indexing (MLSI) [9], Multi-label Dimensionality reduction via Dependence Maximization (MDDM) [10], Multi-Label Least Square (MLLS) [11], Multi-label Linear Discriminant Analysis (MLDA) [12].

The experiment used 5-fold cross validation to evaluate the classification performance of different algorithms when dimension is $K-1$. K-Nearest Neighbour (KNN) classifier is then used after each algorithm. As we can see from Table IV, V) and VI, Multi-label MCDA clearly outperforms other 4 algorithms.

## X. RELATED WORK

Researchers and engineers nowadays have larger and larger data with very high dimension to be processed everyday [1]. Many big data technologies including cloud computing, dimension reduction, accelerating algorithms have been proposed [28] [29] [30] [31] [32] [33]. Trace ratio problem has been studied thoroughly in recent years. Many dimension reduction algorithms can be reduced to a trace ratio objective. But trace ratio problem does not have closed-form solution. Thus how to solve trace ratio efficiently becomes an interesting research topic. Wang [6] proposed an efficient iterative algorithm to get an approximate solution. Shen [34] proposed a formulation for solving the trace ratio problem directly. Nie proposed a Trace Ratio criteria for feature selection[35]. Each feature subset has a feature score, which is computed by trace ratio. They propose an iterative algorithm to find the global optimal feature subset. A number of LDA reformulation ideas have be proposed in recent years, such as PCA+LDA [36], regularized LDA(RLDA) [23], null space LDA (NLDA) [5], Orthogonal Centroid Method (OCM) [25], Uncorrelated LDA(ULDA)[24], Orthogonal LDA (OLDA)[24], etc. Ye introduced a unified framework for generalized LDA in [37]. The unified framework consists of four steps:

1) Compute the eigenvalues $\{\lambda_i\}_{i=1}^d$ and eigenvectors $\{u_i\}_{i=1}^d$ of total covariance matrix $S_t$, where $d$ is the dimension of data. So $S_t = \sum_{i=1}^d \lambda_i u_i u_i^T$.
2) Given a transfer function $\phi$: $\tilde{\lambda}_i = \phi(\lambda_i)$. Construct $\tilde{S}_t = \sum_{i=1}^d \tilde{\lambda}_i u_i u_i^T$.
3) Compute the eigenvectors of matrix $\tilde{S}_t^+ S_b$ that correspond to the largest $q$ eigenvalues, where $q$ is the rank of $S_b$ and $\tilde{S}_t^+$ means pseudo-inverse of $\tilde{S}_t$. Construct matrix $G$ using these $q$ eigenvectors.
4) Optional: compute the QR decomposition of $G = QR$.

The final projection is given as $G$ or $Q$. In RLDA, the transfer function is $\phi(\lambda_i) = \lambda_i + \mu$. In ULDA, $\phi(\lambda_i) = \lambda_i$ and the optional QR decomposition is not applied. In OLDA, $\phi(\lambda_i) = \lambda_i + \mu$ and the optional QR decomposition is applied. In OCM, the optimal transformation is the top eigenvectors of $S_b$ and the transfer function is $\phi(\lambda_i) = 1$.

## XI. CONCLUSIONS

In this paper, we propose a harmonic mean based Linear Discriminant Analysis, Multi-Class Discriminant Analysis (MCDA), which makes use of weighted harmonic mean of pairwise between-class distance and gives higher priority to maximize small between-class distances. MCDA has been extended to multi-label dimension reduction. Extensive experiments on 7 single-label datasets and 4 multi-label datasets show that MCDA outperforms 10 other single-label algorithms and 4 multi-label algorithms consistently in terms of classification accuracy, macro and micro average F1 score.

### ACKNOWLEDGMENT

This work is partially supported by NSF1356628, NSF1633753.